# Multispectral-NeRF：A Multispectral Modeling Approach based on Neural Radiance Fields

Hong Zhang[1,2,3,4], Fei Guo[1,2,3,4], Zihan Xie[1,2,3,4,*], Dizhao Yao[1,2,3,4]

[1]State Key Laboratory of Climate System Prediction and Risk Management, Nanjing Normal University, Nanjing 210023, China;

[2]Key Lab of Virtual Geographic Environment (MOE), Nanjing Normal University, Nanjing 210023, China;

[3]State Key Laboratory Cultivation Base of Geographical Environment Evolution (Jiangsu Province), Nanjing, 210023, China;

[4]Jiangsu Center for Collaborative Innovation in Geographical Information Resource Development and Application, Nanjing, 210023, China

**Abstract:** 3D reconstruction technology generates three-dimensional representations of real-world objects, scenes, or environments using sensor data such as 2D images, with extensive applications in robotics, autonomous vehicles, and virtual reality systems. Traditional 3D reconstruction techniques based on 2D images typically relies on RGB spectral information. With advances in sensor technology, additional spectral bands beyond RGB have been increasingly incorporated into 3D reconstruction workflows. Existing methods that integrate these expanded spectral data often suffer from expensive scheme prices, low accuracy and poor geometric features. Three - dimensional reconstruction based on NeRF can effectively address the various issues in current multispectral 3D reconstruction methods, producing high - precision and high - quality reconstruction results. However, currently, NeRF and some improved models such as NeRFacto are trained on three - band data and cannot take into account the multi - band information. To address this problem, we propose Multispectral-NeRF – an enhanced neural architecture derived from NeRF that can effectively integrates multispectral information. Our technical contributions comprise threefold modifications: Expanding hidden layer dimensionality to accommodate 6-band spectral inputs; Redesigning residual functions to optimize spectral discrepancy calculations between reconstructed and reference images; Adapting data compression modules to address the increased bit-depth requirements of multispectral imagery.

Experimental results confirm that Multispectral-NeRF successfully processes multi-band spectral features while accurately preserving the original scenes' spectral characteristics.

# 1. Introduction

3D reconstruction digitally captures the spatial structure and geometric characteristics of real-world objects or scenes [1,2]. This process fundamentally transforms visual information from sensor data or 2D images into high-precision 3D models through single-view or multi-view analysis [3], enabling accurate digital replication of physical environments. Primary data sources encompass airborne LiDAR, 2D GIS datasets [4], aerial photography, and terrestrial laser scanning [5] or UAV-captured imagery [6]. In the field of 3D reconstruction from 2D images, continuous advancements in sensor technology have enabled multispectral imagery to capture richer spectral feature information compared to traditional true-color (RGB) imagery [7]. When 3D models reconstructed from RGB spectral information are employed for land cover classification and tree species identification, they often exhibit insufficient accuracy. In terms of accurately distinguishing tree species that are morphologically or color-wise similar with minimal physiological differences, as well as identifying vegetation under stress conditions such as pests, drought, or nutrient deficiency, the analysis based solely on visible light data has shown limited performance. Compared to traditional 3D reconstruction methods that rely solely on RGB spectral information, integrating data from multispectral sensors provides a more precise and comprehensive characterization of the underlying scene [8]. Combining this spectral information with 3D geometric data enhances the effectiveness of spatial analysis in three-dimensional space, combining multispectral data can effectively improve tree geometric structure identification and the accuracy of tree species identification in different environments [9]. The additional spectral bands provided by multispectral sensors (such as near-infrared band, red edge band) provide unique spectral information about vegetation that RGB bands lacks, offering greater flexibility in band combinations, enabling more accurate extraction of vegetation structure [10]. This capability provides distinct advantages in analyses such as tree species classification, estimation of forest age and canopy height, above-ground biomass estimation, and the assessment of vegetation growth status. Current multispectral-integrated reconstruction methodologies, however, confront persistent technical challenges: low accuracy, poor geometric features, and being limited by data resolution [11].

With the advancement of deep learning techniques, deep learning-based 3D reconstruction methods have introduced novel solutions to address the limitations of traditional approaches in multispectral 3D reconstruction [12]. Neural Radiance Fields (NeRF)-based 3D reconstruction methods implicitly represent entire scenes by training neural networks, producing high-precision and high-quality realistic outputs [13-15]. NeRF efficiently and effectively generates 3D scenes that integrate both spectral and geometric features, making it particularly suitable for reconstructing multispectral images, which often have lower resolution compared to true-color imagery [7,8].

In conclusion, combining neural radiance fields with multispectral data for 3D reconstruction can maximize the utilization of the rich spectral information in multispectral data. However, the original NeRF framework is specifically designed for three-band data, making it challenging to effectively leverage the advantages offered by multispectral information. Therefore, in this paper, we propose a novel Multispectral-NeRF model based on the NeRFacto framework. This model effectively combines multispectral data to enrich spectral information, addressing various challenges faced by traditional 3D reconstruction methods and existing approaches that incorporate multispectral data, enriching the spectral information input available to the original NeRFacto model, enabling high-precision reconstruction and visualization of 3D scenes with multispectral characteristics. In addition, this paper also discusses the selection of hardware-based model training parameters to ensure that the model achieves optimal performance under different hardware conditions.

## 2. Related Work

## 2.1 Traditional Multispectral-Integrated 3D Reconstruction Methods

Traditional 3D reconstruction typically relies on spectral information from the RGB spectrum. The advancement of sensor technology enables multispectral cameras to capture richer spectral information compared to true-color cameras. Consequently, three-dimensional models derived from multispectral data demonstrate enhanced fidelity in representing real-world conditions, thereby facilitating more effective extraction of actionable insights from spatial datasets [7,8]. Compared with three - band visible light data, multispectral data have a powerful ability to distinguish the physical properties of ground objects thanks to their rich spectral information. For example, some red - edge bands are extremely sensitive to the physiological state of vegetation, which can reflect various biological information of vegetation, identify problems, and be applied in agricultural management. The near - infrared band can be used to detect pollutants in water, the atmosphere, and soil, and it has important applications in ecological environment monitoring [16,17]. Current methodologies for integrating geometric and spectral features in 3D reconstruction can be categorized into three primary approaches: indirect methods, direct methods, and computer vision-based techniques. Indirect methods achieve feature fusion by combining geometrically defined 3D models with spectrally enriched 2D imagery. However, this approach exhibits significant dependence on precise georegistration accuracy, with final model quality being disproportionately affected by registration errors [18]. Direct methods employ multispectral LiDAR systems to simultaneously capture geometric and spectral attributes through terrestrial scanning. Nevertheless, conventional implementations typically limit spectral resolution to true-color or predetermined bands, while maintaining separate acquisition workflows for geometric and spectral data [19,20]. Computer vision-based techniques predominantly rely on traditional Multi-View Stereo (MVS) image matching. The process generally involves feature detection and matching, sparse point cloud generation, estimation of camera intrinsic and extrinsic parameters, bundle adjustment, and dense reconstruction to build the final 3D model, yet encounter persistent challenges including insufficient point cloud density, surface discontinuities, and residual overlapping artifacts during photogrammetric point cloud generation [11]. While commercial software packages implementing computer vision algorithms can generate multispectral-colored point clouds via MVS processing, the inherent resolution limitations of multispectral imagery compared to true-color counterparts result in suboptimal point cloud densities. This deficiency becomes particularly pronounced in occluded areas with insufficient photo coverage, such as beneath vegetation canopies, where inadequate feature point extraction leads to geometric reconstruction failures and surface voids [11,21,22]. As a 3D reconstruction algorithm in the field of machine learning, NeRF can effectively address the issues existing in the aforementioned methods.

## 2.2 NeRF-Based 3D Reconstruction Methods

The Neural Radiance Field (NeRF) model [23] presents an innovative view synthesis approach that combines volume rendering with implicit neural scene representations. By training neural networks to generate high-fidelity 3D reconstructions, this method effectively mitigates common issues in traditional approaches such as sparse point clouds and rendering artifacts. This technique reconstructs 3D radiance fields by training neural networks to model scene geometry and illumination characteristics using multi-view 2D images and associated camera parameters as input. As illustrated in **Figure 1**, NeRF's core innovation lies in employing Multilayer Perceptron (MLP) networks to continuously model volumetric density and color distributions. Through differentiable volume rendering, the system generates view-dependent color and density values for any spatial coordinate and viewing direction. Recent years have witnessed significant advancements in the field of NeRF. Xu et al. integrated the advantages of NeRF and MVS by embedding feature-rich point clouds into the volumetric rendering pipeline. Specifically, they utilized neural 3D point clouds with learned neural features to approximate the radiance field [24]. Furthermore, they introduced point cloud pruning and generation algorithms to accelerate both reconstruction and rendering, while simultaneously improving the quality of the generated point clouds [25]. Chen et al. proposed a mesh-based NeRF-like model, in which they assigned color, feature, and opacity MLPs to each mesh element. Additionally, they employed a lightweight MLP to handle pixel-wise shading, which significantly improved the training efficiency [26]. Muller et al. introduced learnable multi-resolution hash encodings, which were jointly optimized with the NeRF model and MLP parameters. This approach substantially enhanced both training and inference speeds, while also improving the accuracy of scene reconstruction using NeRF-based models [27]. These studies demonstrate significant advancements in terms of reconstruction fidelity enhancing, training and inference acceleration.

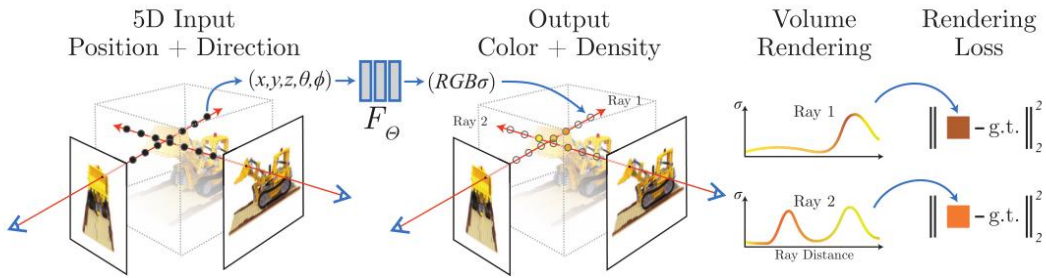

**Figure 1.** Workflow of NeRF Model [23]

Building upon Mip-NeRF-360, the NeRFacto model [28] integrates multiple NeRF enhancement techniques (**Figure 2**) that optimize camera pose estimation, sampling strategy, and spatial encoding. The framework employs a piecewise sampler to generate initial scene samples, allocating 50% of samples uniformly within the camera's near-clip range. Subsequent refinement is achieved through a dual-process system: the proposal sampler concentrates samples in regions contributing most significantly to final rendering, while the density field utilizes coarse scene representations to guide sample distribution [30-32]. The model enhances view synthesis through optimized SE transformations [32,33], which are processed by Mip-NeRF-360's proposal network sampler. Scene density is computed via a compact fused MLP with multi-resolution hash encoding [30], enabling effective 3D reconstruction in unbounded environments. Compared to Instant-NGP's rapid generation approach, while slightly slower in training speed, NeRFacto demonstrates enhanced performance in handling complex lighting conditions and surface textures characteristic of natural environments. NeRFacto maintains rendering fidelity while significantly improving training efficiency, these advancements establish NeRFacto as an optimal foundation for our multispectral 3D reconstruction research.

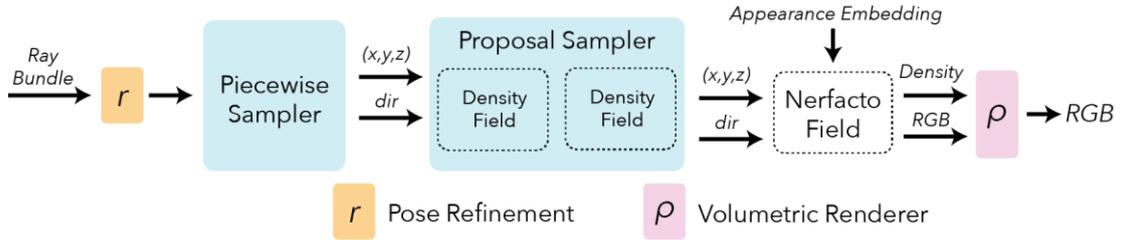

**Figure 2.** Workflow of NeRFacto Model [28]

Current research predominantly employs visible-light data for NeRF-based 3D reconstruction and point cloud generation, while few studies have explored integrating Neural Radiance Fields (NeRF) with multispectral data to enrich spectral information. Chen et al. decomposed a 4D scene tensor into multiple compact low-rank tensor components for neural radiance field rendering. In their method, each voxel contains a 4D tensor with multi-channel features. While this approach supports multi-band processing, it focuses exclusively on true-color bands and overlooks challenges specific to multispectral image modeling [31]. The Spec-NeRF model proposed by Li et al. is based on the original NeRF framework for scene reconstruction. However, its underlying mechanism still relies on images generated through filters by true-color cameras. The method focuses on reconstructing multispectral images and is primarily oriented toward synthesizing multispectral image datasets based on spectral curves [34].

Matteo et al. introduced X-NeRF, which can simulate scenes from images across different spectral bands. However, the spectral information originates from multiple camera payloads, making it challenging to compute the relative poses between them [35]. As an implicit representation, NeRF excels at reconstructing models with complex topological structures and overcomes resolution constraints. This capability makes it particularly suitable for reconstructing multispectral imagery, which often exhibits lower spatial resolution compared to true-color images [36]. Unlike LiDAR-based point cloud acquisition methods that focus solely on geometric features, NeRF's implicit scene representation enables simultaneous extraction of both geometric attributes and spectral characteristics. However, both the original NeRF model and the improved NeRFacto model are designed to handle standard three - band data. Therefore, this paper improves the NeRFacto model to address this issue.

3D Gaussian Splatting [37] is a real-time radiance field rendering technique that represents scenes using 3D Gaussian ellipsoids. By leveraging specialized optimization and rendering algorithms, it achieves high-quality, real-time novel view synthesis. Both 3D Gaussian Splatting and NeRF are prominent techniques in the field of novel view synthesis [38,39,40], aiming to generate high-fidelity images from arbitrary viewpoints and delivering strong performance in 3D reconstruction. However, they are not entirely interchangeable; their data outputs and application domains differ, making each more suitable for specific use cases [41]. Although the output of 3D Gaussian Splatting resembles point cloud data, it essentially represents the spatial centers of 3D Gaussian ellipsoids rather than actual point clouds. In contrast, NeRF models can generate true point clouds through inference, which provides a distinct advantage in tasks that require integrating geometric and spectral features. While 3D Gaussian Splatting excels in rendering efficiency and computational cost, NeRF models demonstrate superior performance in rendering quality [42]. Furthermore, improvements to the implicit representation of NeRF enable advanced volumetric operations [43,44], such as volumetric space construction and individual tree segmentation in vegetation analysis—capabilities that 3D Gaussian Splatting, as an explicit representation method [45], cannot support. Therefore, between the two popular techniques in novel view synthesis—3D Gaussian Splatting and NeRF—we have opted to employ NeRF to achieve our research objectives.

## 3. Methodology

## 3.1 Model Improvements

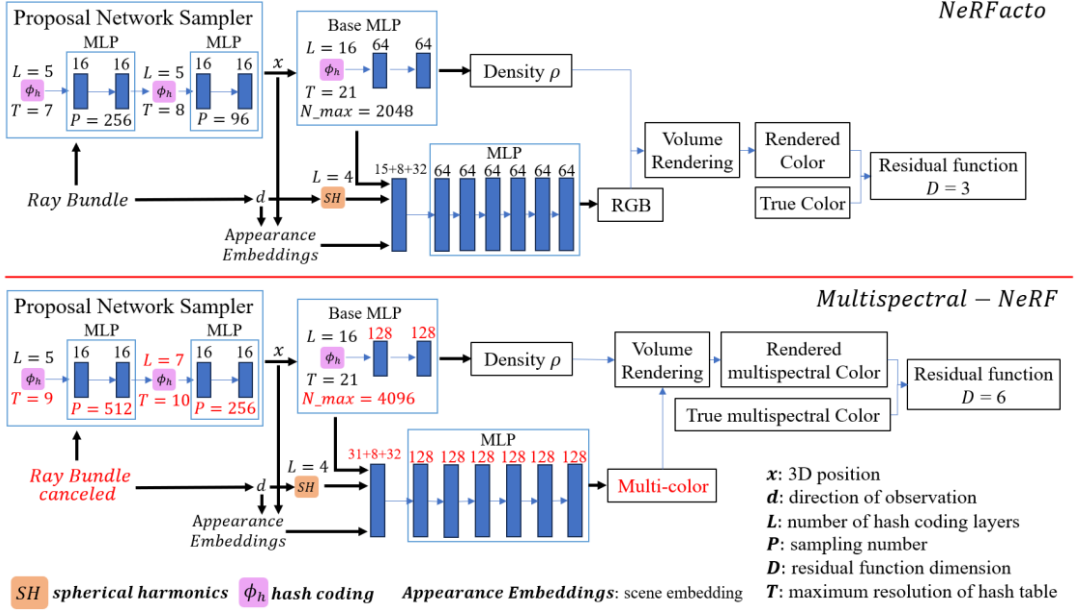

**Figure 3.** Workflow comparison of NeRFacto [28] and Multispectral-NeRF

To address the limitation of NeRFacto's exclusive reliance on RGB imagery (both authentic and synthetic) for 3D reconstruction, this study proposes a modified framework that synergizes six-band multispectral sensor data with NeRFacto's architecture, thereby expanding its spectral discriminative capacity.

To address the higher bit-depth characteristics of multispectral imagery compared to standard RGB images, we modified the ray input component in the training set (**Figure 4**). Unlike conventional 8-bit or 10-bit RGB images, multispectral data typically exhibit 16-bit or 32-bit depth, encompassing broader pixel value ranges that capture finer radiometric distinctions. The original NeRFacto framework linearly compresses RGB pixel values (0-255 range) to 0-1 during preprocessing, subsequently restoring them post-training. This normalization approach fundamentally alters data bit-depth, thereby discarding critical radiometric information inherent in multispectral datasets. In contrast, preprocessed multispectral pixels (after band registration and radiometric calibration) directly represent surface reflectance as 0-1 floating-point values, inherently bypassing the need for linear compression. We revised the linear compression function in NeRFacto's ray input module to preserve native bit-depth during residual calculations and backpropagation, effectively maintaining spectral fidelity throughout the learning process.

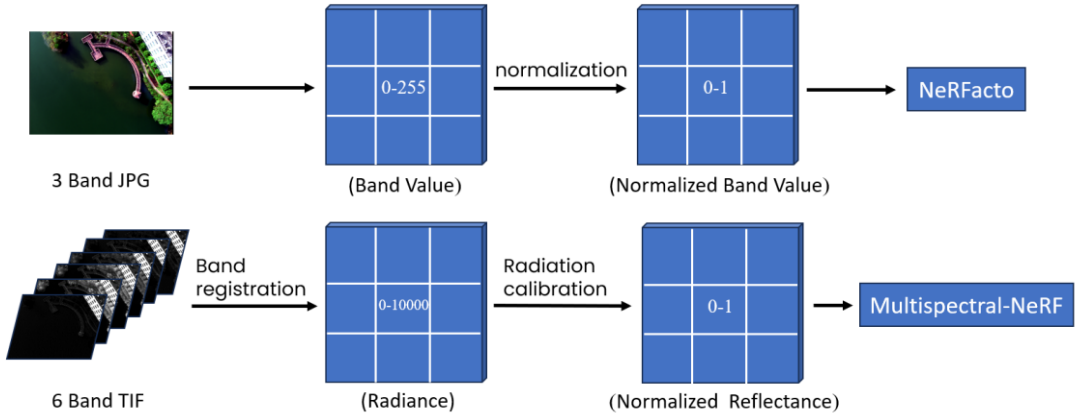

**Figure 4.** Data Depth Adaptation

Furthermore, to integrate spectral information from additional bands, we extended the original NeRFacto model's primary MLP neural network output from 3-band to 6-band multispectral color results through parameter optimization in the spectral network's hidden layers. This modification enables simultaneous processing of all bands in multispectral imagery. Drawing on the grid search hyperparameter optimization approach from Instant-NGP [30], we refined the color hidden layer dimensions of the primary MLP network to enhance characterization accuracy of multispectral features. We expanded both the color hidden layer dimensions in the main MLP and the base MLP's hidden layers (**Figure 5**). Concurrent adjustments were made to the output geospatial feature vector dimensions, main MLP input dimensions, and subsequent color rendering functions to ensure full architectural compatibility.

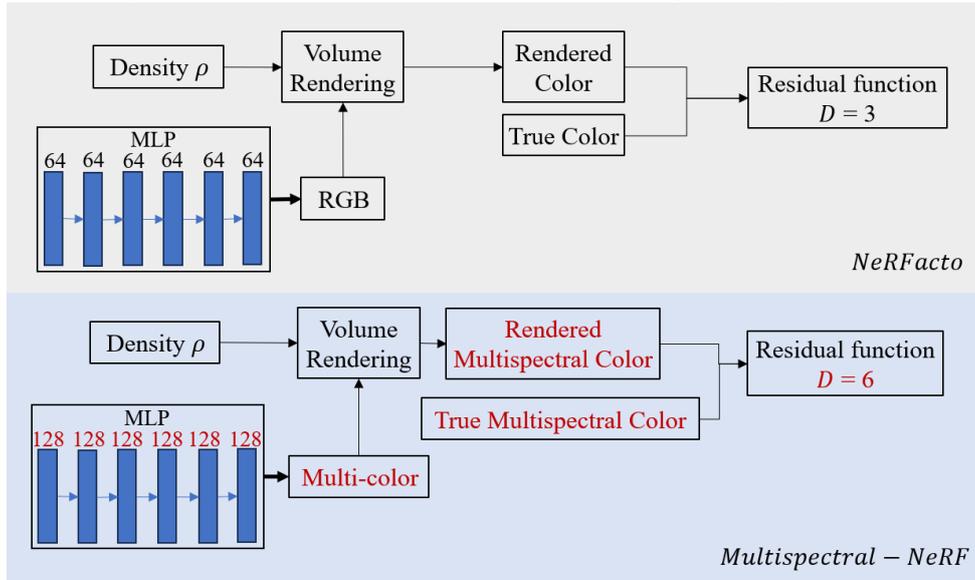

**Figure 5.** Spectral band expansion comparison of NeRFacto [28] and Multispectral-NeRF

To address the limitation of NeRFacto's tri-band residual function in processing multispectral data, we redesigned its residual computation module. The refined framework minimizes discrepancies between predicted and ground-truth images by calculating the mean squared error (MSE) between reconstructed and original spectral values across all bands, as formalized below:

$$L = \frac{1}{NB} \sum_{p=1}^{N} \sum_{b=1}^{B} (I_{p,b} - \hat{I}_{p,b})^2 \qquad (3.1)$$

Where $I_{p,b}$ denotes the ground-truth value at pixel $p$ in band $b$, $\hat{I}_{p,b}$ represents the predicted spectral value, $B$ indicates the total number of spectral bands ($B$=6 in our multi-spectral configuration), and $N$ corresponds to the total pixel count in the image ($N$=1280×960=1228800 in our multi-spectral image). This enhanced loss function accounts for full spectral fidelity across all pixels, enabling comprehensive scene representation optimization through six-band residual analysis between reconstructed and reference multi-spectral imagery.

The enhanced Multispectral-NeRF framework demonstrates optimized processing capabilities for multi-band spectral radiance values, enabling comprehensive multispectral point cloud generation. The architecture utilizes camera intrinsics, extrinsics, and tilted multispectral imagery as inputs for concurrent spatial and feature encoding through dedicated MLP networks. These parallel computational pathways subsequently converge in a central MLP module to derive precise color and density estimates at sampled spatial coordinates. Through volume rendering principles, the system synthesizes predicted multispectral color values that undergo residual analysis against ground-truth spectral data. This differential computation drives backpropagation optimization, progressively minimizing spectral discrepancies while implicitly reconstructing the multispectral representation of the entire scene. The framework innovatively integrates spectral-geometric fusion through depth-aware accumulation mechanisms, enabling simultaneous recovery of both spectral signatures and geometric structures. This dual reconstruction capability culminates in a unified 3D model output that preserves photometric-spectral consistency across all spatial dimensions.

## 3.2 Model Parameter Selection

Given the unique requirements of multispectral data and scene complexity, this study calibrates and optimizes parameters within the compact fusion MLP module of the Multispectral-NeRF framework. These adjustments specifically address the characteristics of experimental zone scenarios and multispectral imaging features, significantly enhancing the model's capacity to process intricate geographic data structures. **Table 1** delineates critical parameter configurations implemented in our Multispectral-NeRF architecture:

Table 1. Parameter selections of NeRFacto [28] and Multispectral-NeRF model

| Parameters | NeRFacto | Multispectral-NeRF |
| --- | --- | --- |
| Hidden Layer Dimension | 64 | 128 |
| Spectral Network Hidden Layer Dimension | 64 | 128 |
| Base MLP Hash Table Max Resolution | 2048 | 4096 |
| Base MLP Hash Table Size | $2^{19}$ | $2^{21}$ |
| Base MLP Output Geometric Feature Dimension | 15 | 31 |
| Proposal Network Sampler Sample Points | 256,96 | 512, 256 |
| Proposal Network Sampler Hash Table Max Resolution | 128,256 | 512,1024 |
| Proposal Network Sampler Hash Table Max Levels | 5,5 | 5,7 |

The model parameter selection comprises three key components: hidden layer dimensionality enhancement based on multispectral and multichannel data, multi-resolution hash encoding refinement, and proposal network sampler parameter adjustments.

First, the hidden layer dimensionality of both the spectral network and main architecture in Multispectral-NeRF was standardized at 128. This adjustment addresses the complex spectral characteristics of multispectral imagery, which contains 6-12 spectral bands compared to conventional RGB's 3 channels. The 128-dimensional configuration balances feature learning capacity for intricate geospatial patterns with computational efficiency, enabling high-dimensional feature extraction for precise volumetric rendering while maintaining operational viability with large datasets.

Secondly, the hash table's maximum resolution was elevated to 4096 to accommodate large-scale geospatial modeling requirements. This enhancement permits finer spatial detail resolution, crucial for capturing subtle spectral variations across expansive terrain. To maintain computational tractability, the implementation incorporates an optimized voxel hierarchy structure within the multi-resolution hash encoding framework.

Thirdly, the proposal network sampler was augmented with increased sampling points during the secondary sampling phase, enhancing spatial-spectral information acquisition for improved depth estimation and 3D reconstruction fidelity. While retaining a compact fused MLP architecture, the sampler's hash resolution was proportionally scaled with additional network layers to resolve fine-grained environmental details without compromising inference speed.

The method for determining the parameters refers to the grid search hyperparameter tuning method in InstaNGP [30]. Taking the MLP as an example, experiments were conducted on the relationship between training time and training error when the dimensionality of the color hidden layer of the model was extended to different values. The results are shown in **Figure 6**. Under different numbers of hidden layers, when the dimensionality of the hidden layer was extended to 128, high - precision results were obtained in a relatively short training time. Therefore, the dimensionality of the color hidden layer of the overall MLP and the hidden layer of the Base MLP in the Multispectral - NeRF model were extended to 12.

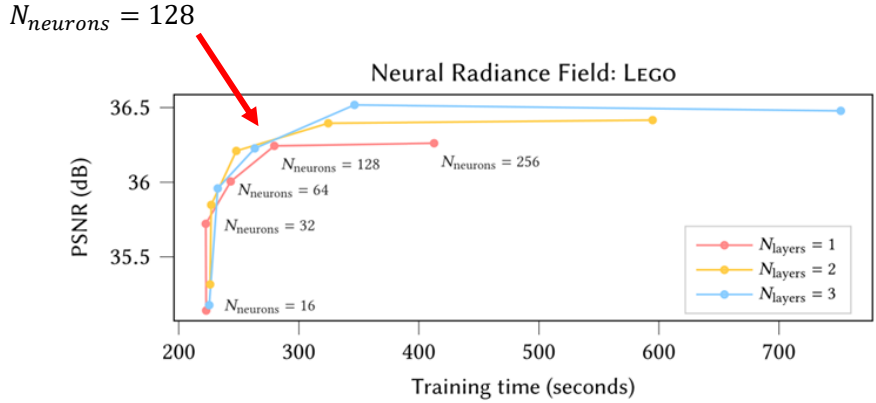

**Figure 6.** MLP Hyperparameter Tuning Results [30]

Collectively, these parameter optimizations enable the Multispectral-NeRF framework to process 6-band UAV-acquired multispectral imagery from MS600Pro sensors with enhanced geometric and spectral reconstruction capabilities. The architectural refinements specifically improve feature disentanglement across spatial, geometric, and spectral domains, achieving superior reconstruction accuracy for complex real-world environments.

## 4. Experiments and Data

### 4.1 Study Area

In this study, we utilized the School of Geographic Sciences located in the northern part of Nanjing Normal University's Xianlin Campus as our experimental area. Nanjing Normal University's Xianlin Campus is situated in Qixia District, Nanjing City, Jiangsu Province, with geographic coordinates approximately at 118.910321°E and 32.105220°N. It lies on the eastern foot of Zijin Mountain in western Qixia District, with altitudes generally ranging from 20 to 40 meters, aside from several small hills within the campus grounds. Twelve ground control points (GCPs) were strategically distributed across the study area and measured using RTK positioning via the Huace i93 GNSS receiver. The distribution of these control points is illustrated in **Figure 7**.

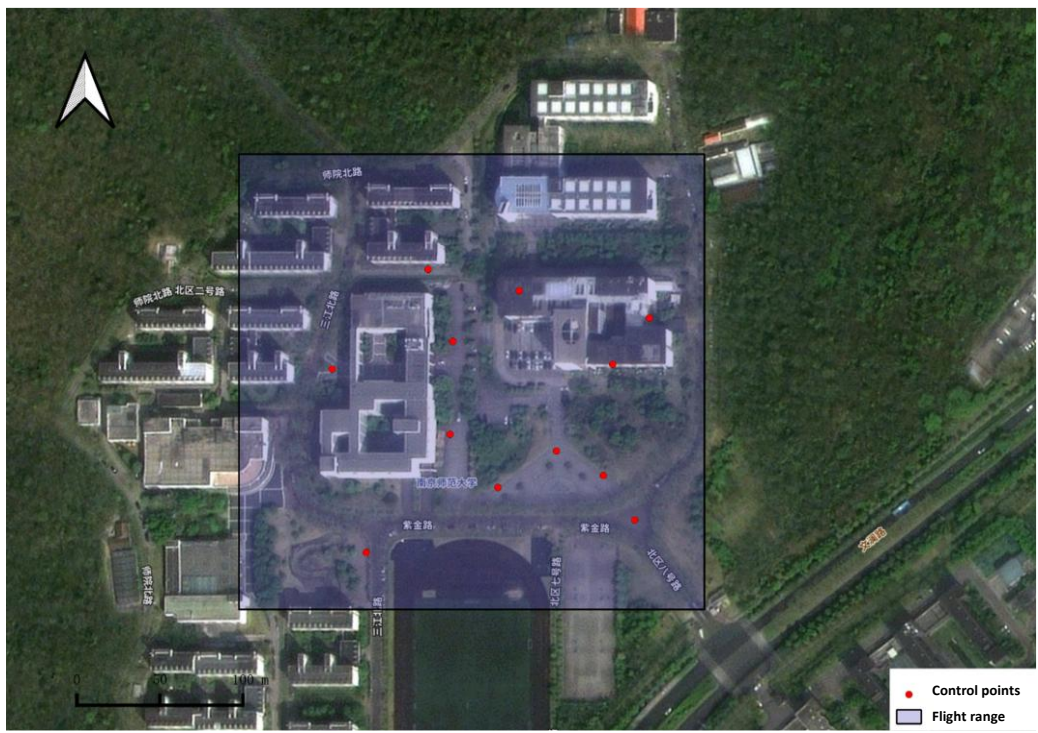

**Figure 7.** Layout of GCPs in the experimental area

## 4.2 Experimental Setup

In this study, we employed the DJI M300RTK multi-rotor drone (**Figure 8**(a)) as our aerial platform. We utilized the MS600Pro multispectral sensor (**Figure 8**(b)) to capture multispectral data of the study area, which was then used to generate a multispectral point cloud. Additionally, the DJI Zenmuse L1 payload (**Figure 8**(a)) was deployed to acquire LiDAR point cloud data for validating the accuracy of the multispectral point cloud. Data collection was facilitated by the DJI D-RTK2 mobile station serving as our RTK positioning system, receiving signals from BeiDou satellites to provide real-time differential positioning data, thereby achieving centimeter-level accuracy for the drone. Detailed specifications of the MS600Pro multispectral camera are summarized in **Table 2**.

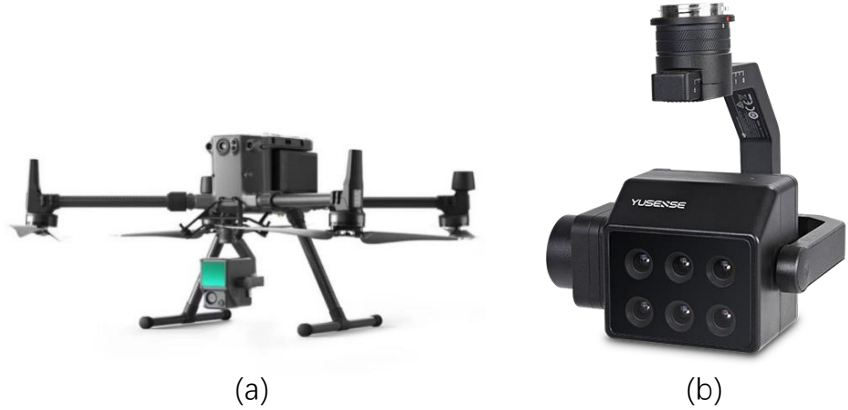

(a)                  (b)

**Figure 8.** DJI M300RTK with L1(a) and MS600pro(b)

**Table 2.** Detailed parameters of MS600Pro

| Technical parameters | Value |
| --- | --- |
| Effective Pixel of Sensor | 1.2 million pixels |
| Quantization Bit | 12bit |
| Field of View | HFOV：49.6° <br> VFOV：38° |
| Focal Length | 5.2mm |
| Ground Sample Distance | 8.65cm@h=120m <br> translation：±320° |
| Spectral Band Range | B:450nm, G:555nm, R:660nm, RE1:720nm, RE2:750nm, NIR:840nm |
| Swath Width | 110m*83m@h=120m |

During data collection using drones equipped with various payloads, the circling flight paths capture images from multiple angles, providing significant advantages for data acquisition over buildings, landscapes, landmarks, etc. Compared to nadir flight paths, fewer images are needed to cover all perspectives more comprehensively, effectively capturing all angles and directions of surface features. In addition, the circular flight path can improve data collection efficiency, reducing both flight frequency and time costs.

To accommodate terrain variations in the experimental area, terrain-following adjustments were integrated into the orbital flight planning to ensure adaptive altitude changes according to terrain. The parameters of the terrain - following circling flight path used in **Table 3**, and the schematic diagram of the terrain - following circular flight path planning result is shown in **Figure 9**. Data were collected in September 2023 under conditions of clear skies, abundant sunlight, and wind speeds below 1 m/s.

**Table 3.** Circling flight path parameter settings

| Route parameter | Value |
|---|---|
| Relative Flying Height | 100m |
| Inter-Circumaviation Overlap Rate | 75% |
| Intra-Circumaviation Overlap Rate | 75% |
| Gimbal Tilt Angle | -60° |
| Boundary Buffer | 15m |

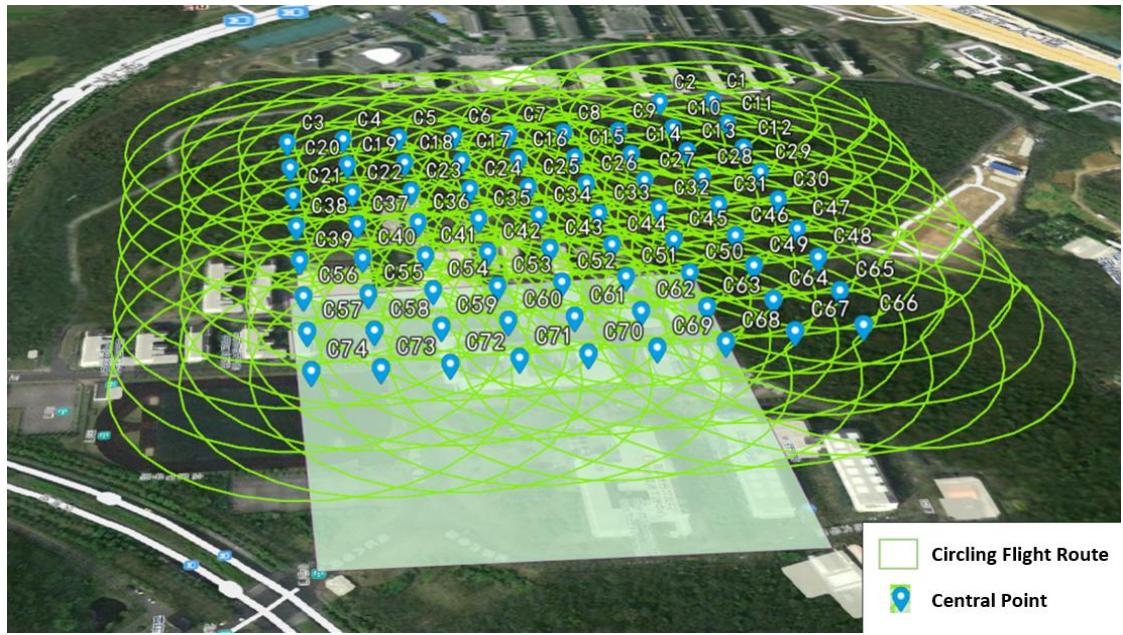

**Figure 9.** Flight control network of study area

## 4.3 Experimental Data Acquisition and Processing

The technical route of this paper is shown in **Figure 10**. First, data from the experimental area is collected. After flight path planning and image control point layout, an aerial survey UAV equipped with sensor payloads is used to obtain oblique photography data. Then, the collected multispectral image data and LiDAR point cloud data are pre - processed separately. For the multispectral image data, medium - precision camera alignment is performed, followed by a second alignment after marking image control points to finally obtain a multispectral dense point cloud. For the L1 LiDAR point cloud data, 3D point cloud reconstruction is carried out. After selecting corresponding points of image control points and calculating the transformation matrix, point cloud image control point registration is performed. Finally, the accuracy comparison is made between the registered LiDAR point cloud data and the multispectral dense point cloud reconstructed from the multispectral data.

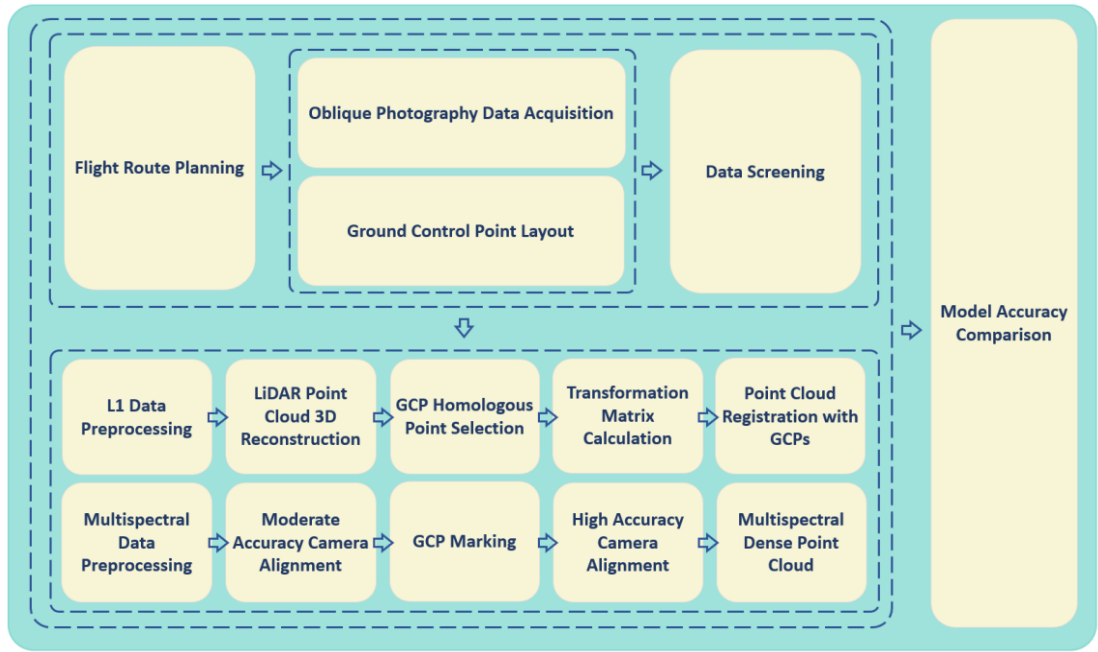

**Figure 10.** Workflow of experimental data acquisition and preprocessing

The multispectral imagery data and L1 LiDAR point cloud data collected in the study area using the data acquisition equipment described in this paper are summarized in **Table 4**:

**Table 4.** Experimental data

| Data type | Number of photographs | Data size |
| --- | --- | --- |
| Multispectral images | 3171 | 92.9GB |
| Lidar point cloud | / | 15.7GB |

The preprocessing of multispectral raw data involves two key steps: band registration and radiometric calibration. Each dataset comprises six single-band images, with ten groups exhibiting distinct reflectance features selected for interior orientation. This process projects images onto a pre-calibrated coordinate system to determine fiducial point coordinates, utilizing these high-confidence references to build a transformation model via affine transformation or polynomial fitting, thereby mapping arbitrary image points to the system. Subsequently, band registration for the UAV-based multispectral imagery's six bands is conducted using the interior orientation outcomes. This correction enhances geometric accuracy by eliminating image deformations, establishing a reliable foundation for subsequent data processing and analysis. Radiometric calibration is the process of determining and calibrating the relationship between digital values in remote sensing im-ages and physical radiation quantities using experiments or standard reference sources. In remote sensing data processing and analysis, this calibration converts digital values into physical radiation quantities, such as radiance or reflectance, enabling quantitative analysis and study of surface features. A gray card serves as a standard reference source with known radiation characteristics, which can be precisely determined through experiments and measurements. By imaging the gray card across six spectral bands, the reflectance for each band is obtained and used as the basis for radiometric calibration. Consequently, after calibration, the images record the absolute reflectance of ground objects, accurately reflecting their spectral reflection properties across the electromagnetic spectrum; this data facilitates the calculation of various indices commonly used in remote sensing image interpretation. During calibration for each band, radiance is converted to reflectance by incorporating calculations based on the calibrated gray card.

After the preprocessing steps, a 3D model is reconstructed. Ground control points are used for secondary point picking to improve the accuracy, and finally, a multi - spectral dense point cloud with six bands is generated. As shown in **Figure 11**, it is the aerial triangulation quality report generated after the GCP correction, which is achieved through secondary point picking based on GCPs. Among them, the Z errors are represented by the colors of the ellipses, while the X and Y errors are represented by the shapes of the ellipses. The estimated positions of GCPs are represented by dots or crosses, and the pixel errors are controlled within 0.5 pixels to ensure accuracy. The error accuracy of GCPs in the experimental area is shown in **Table 5**.

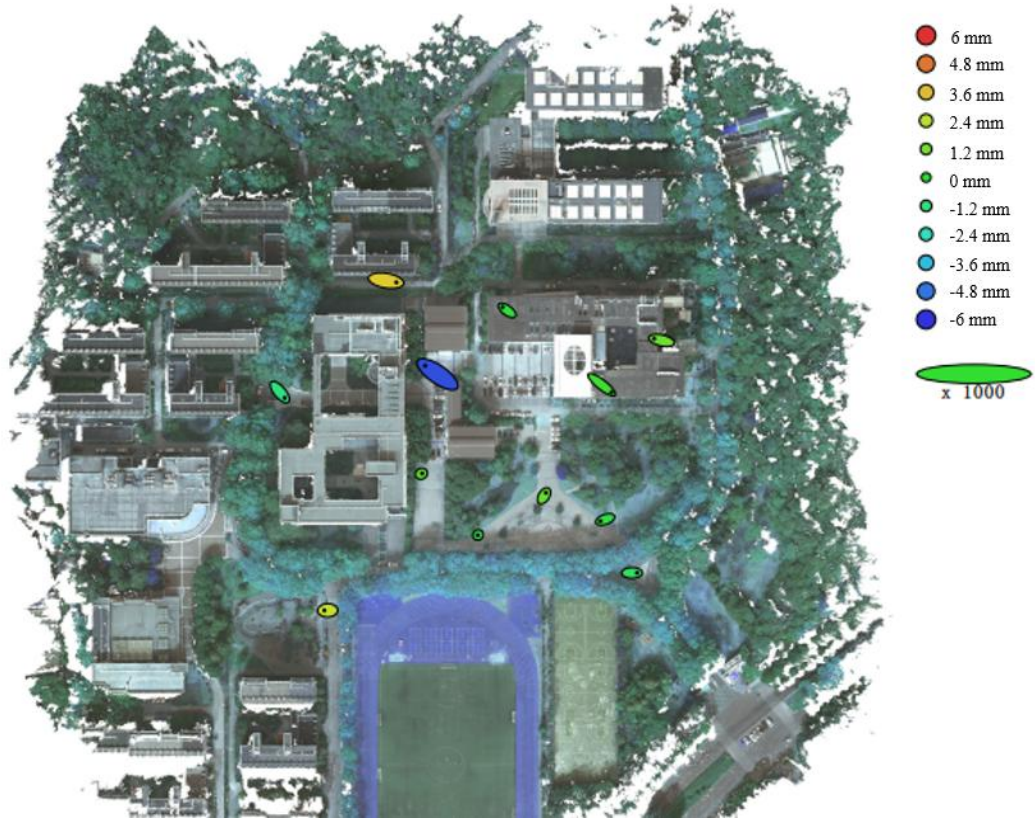

**Figure 11.** Repricking of ground control point

**Table 5.** GCP pricking accuracy report

| Number of GCP | X Error(mm) | Y Error (mm) | Z Error (mm) | Aggregate (mm) | Image pixel error (pi) | Projected quantity |
|---|---|---|---|---|---|---|
| 1 | 10.03 | -7.95 | 0.23 | 12.80 | 0.38 | 41 |
| 2 | -7.70 | 1.77 | 1.06 | 7.98 | 0.39 | 28 |
| 3 | -4.68 | 3.38 | -0.19 | 5.78 | 0.39 | 17 |
| 4 | 1.95 | 3.20 | 0.84 | 3.85 | 0.41 | 30 |
| 5 | -0.27 | -0.21 | -0.51 | 0.61 | 0.42 | 26 |
| 6 | -4.84 | -1.76 | -0.38 | 5.16 | 0.43 | 21 |
| 7 | 5.17 | 0.03 | -0.87 | 5.24 | 0.33 | 25 |
| 8 | 0.78 | 0.37 | 0.82 | 1.19 | 0.38 | 16 |
| 9 | -12.82 | 8.92 | -5.49 | 16.56 | 0.44 | 61 |
| 10 | 10.76 | -1.78 | 3.34 | 11.41 | 0.44 | 36 |
| 11 | 5.00 | -5.89 | -1.37 | 7.85 | 0.44 | 25 |
| 12 | -3.39 | -0.07 | 2.57 | 4.26 | 0.40 | 41 |
| Average | 6.79 | 4.17 | 2.11 | 8.24 | 0.41 | 31 |

Finally, after marking the image control points, the pixel error of the experiment was controlled within 0.5 pixels. Then, high - precision aerial triangulation was performed again to calculate the camera pose, and a depth map was generated to estimate the point cloud model (**Figure 12**).

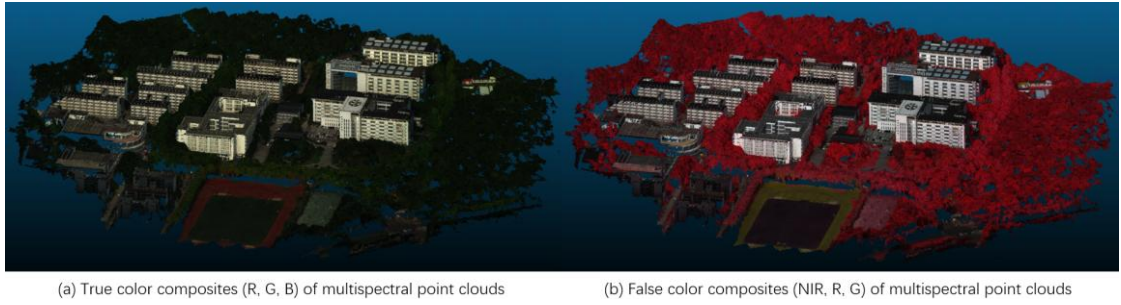

(a) True color composites (R, G, B) of multispectral point clouds   (b) False color composites (NIR, R, G) of multispectral point clouds

**Figure 12.** Multispectral point cloud of study area.

## 4.4 Data Feasibility Analysis

The lidar point cloud generated by the drone L1 is registered with ground control points to serve as the reference data. By comparing the lidar point cloud with the 3D point cloud model generated based on the SFM algorithm, the ECEF (Earth-Centered, Earth-Fixed) coordinate system is used as the spatial reference for the comparative analysis of the Euclidean distance accuracy to assess the geographical coordinate accuracy of the comparative data. The UTM Zone 50N coordinate system was employed as the spatial reference to conduct a comparative analysis of planar and elevation accuracy in geographic coordinates, thereby assessing the accuracy of the geographic coordinate data.

Select the coordinates of 10 corner points of the L1 3D model as the true values and compare them one - by - one with the corresponding points of the 3D model to obtain the accuracy of the primitive points, which reflects the complexity of real - world objects. The geographical coordinate error analysis of the primitive point coordinates is conducted through a combination of the Euclidean Distance, planar error, and elevation error. The Euclidean Distance represents the distance between two points in a 3D coordinate system. The formula for the 3D geographical coordinate accuracy error is as follows:

$$ED(x,y) = \sqrt{\sum_{i=1}^{n}(x_i - y_i)^2} \quad (4.1)$$

Where, $n$ represents the number of dimensions, $x_i$ and $y_i$ represent the components of the corner coordinates of L1 and the corner coordinates of the point - cloud model in the $i$ - th dimension, respectively. By comparing and analyzing the differences in the geographical coordinates of the multi - spectral point - cloud data generated by the SFM algorithm before and after the image control point registration of the L1 lidar, the feasibility of the experimental data in this study is verified. For the feasibility analysis of the data in this study, 10 test points, as shown in **Figure 13**, are selected. These points are evenly distributed across various parts of the experimental area for data feasibility analysis, including building corners, ground corners, and clearly distinguishable points. Coordinates are picked from these points to conduct a feasibility analysis of the geographical coordinate accuracy.

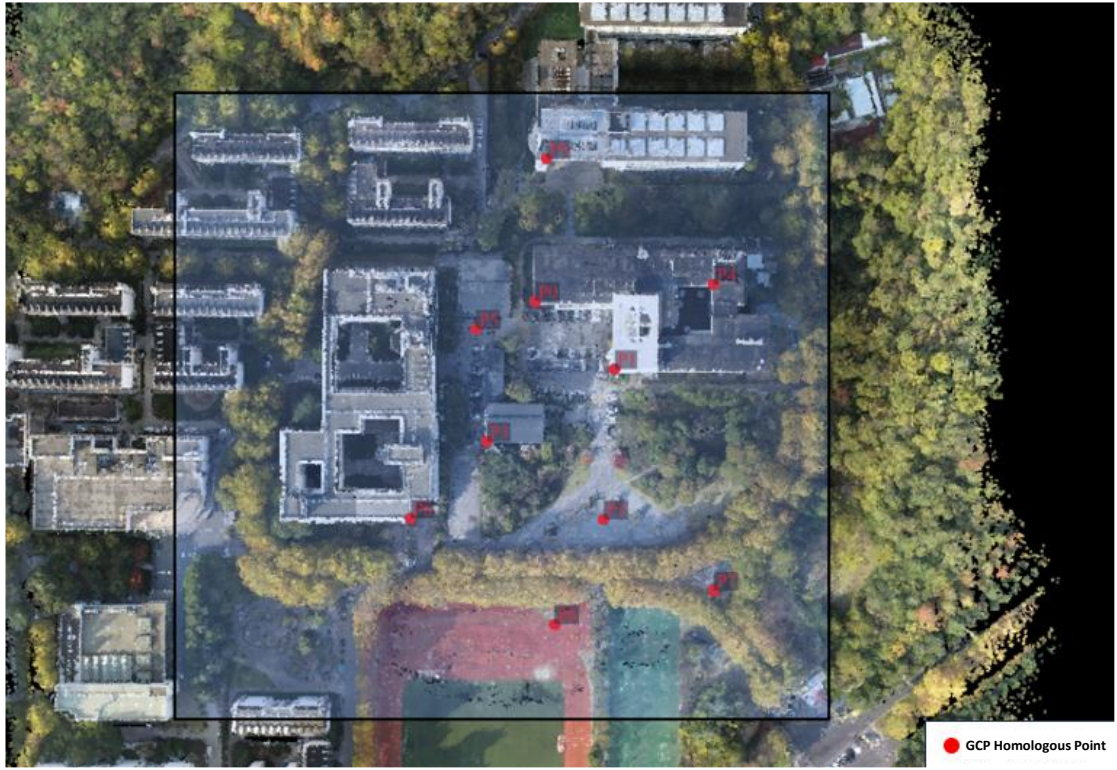

**Figure 13.** Accuracy verification test point

Using the L1 LiDAR point cloud registered with ground control points as the reference point cloud, the Euclidean distance errors, planar errors, and elevation errors between the original reconstructed L1 LiDAR point cloud and the multi - spectral point cloud generated by the SFM algorithm after marking ground control points are shown in **Table 6** and **Table 7**.

**Table 6.** Average Euclidean distance error

| Point cloud category | Mean Euclidean distance error (m) |
| --- | --- |
| L1 Raw Unregistered Point Cloud | 0.178 |
| Multispectral Point Cloud via SFM Algorithm | 0.241 |

**Table 7.** Plane and elevation accuracy error

| Point cloud category | Mean plane distance error (m) | Mean elevation distance error (m) |
| --- | --- | --- |
| L1 Raw Unregistered Point Cloud | 0.196 | 0.092 |
| Multispectral Point Cloud via SFM Algorithm | 0.261 | 0.123 |

Analysis of the results shows that although there are some errors in the point cloud generated by the SFM algorithm compared with the uncorrected L1 LiDAR point cloud, the average Euclidean distance error is within a reasonable range. The multi - spectral point cloud model generated based on the SFM algorithm is feasible for restoring the geographical coordinates in the experimental area and meets the accuracy standard criteria with planimetric accuracy < 0.3 m and vertical accuracy < 0.5 m [46]. Moreover, the multi - spectral point cloud generated by the model has a relatively consistent spatial distribution with the standard reference L1 LiDAR point cloud.

## 5. Result

### 5.1 Accuracy Evaluation Metrics

In the field of 3D reconstruction, the accuracy assessment of models is a crucial step, which directly affects the effectiveness and reliability of model applications. To evaluate the model training accuracy of Multispectral - NeRF based on oblique multispectral photography, this paper conducts model training accuracy evaluation using three key evaluation metrics: Peak Signal - to - Noise Ratio (PSNR), Structural Similarity Index (SSIM), and Learned Perceptual Image Patch Similarity (LPIPS).

The Peak Signal-to-Noise Ratio (PSNR) is a classic metric for evaluating image quality. It assesses based on the ratio between the maximum possible power of the signal and the noise power that affects its quality. In Multispectral-NeRF model, PSNR is used to evaluate the similarity between the reconstructed image and the original image. PSNR is defined based on the Mean Squared Error (MSE), which can be defined as:

$$MSE = \frac{1}{mn} \sum_{i=0}^{m-1} \sum_{j=0}^{n-1} [I(i,j) - K(i,j)]^2 \tag{4.1}$$

The formula for calculating PSNR is as follows:

$$PSNR = 20 \cdot log_{10} \left( \frac{MAX_I}{\sqrt{MSE}} \right) \tag{4.2}$$

Here, $I$ and $K$ are the pixel values of the original image and the reconstructed image respectively, and $m$ and $n$ are the dimensions of the image. $MAX_I$ is the maximum possible pixel value of the image. A higher PSNR value indicates a smaller difference between the reconstructed image and the original image.

The Structural Similarity Index (SSIM) is a more advanced metric used to measure the structural similarity between two images. Unlike PSNR, SSIM not only considers the difference in pixel intensities but also takes into account the structural information of the images. SSIM mainly assesses three key features of an image: Luminance, Contrast, and Structure, which are closer to the perceptual characteristics of the human visual system. The calculation formula for SSIM is:

$$SSIM(x, y) = [l(x, y)^\alpha \cdot c(x, y)^\beta \cdot s(x, y)^\gamma] \qquad (4.3)$$

Here, $l(x, y)^\alpha$ represents the comparison of luminance features, $c(x, y)^\beta$ represents the comparison of contrast features, and $s(x, y)^\gamma$ represents the comparison of structural features, where $\alpha > 0$, $\beta > 0$, $\gamma > 0$. In Multispectral-NeRF model, SSIM is used to evaluate the visual quality and structural fidelity of the reconstructed images.

Learned Perceptual Image Patch Similarity (LPIPS) is an evaluation criterion for assessing the perceptual similarity between a reconstructed image and an original image. Different from traditional pixel - difference - based methods, LPIPS takes into account the high - level features and perceptual characteristics of image content. The core of LPIPS is to extract feature representations of images using a deep - learning model and then calculate the differences between images based on these high - level features. This metric is closer to human visual perception, considering not only the differences in pixel values between images but also the image content, context, and texture. The formula is as follows:

$$LPIPS(x, y) = \Sigma_l w_L \cdot d^l(x, y) \qquad (4.4)$$

Here, $w_l$ represents the weight of the features in the $l$-th layer. These weights can be obtained through training to ensure that the LPIPS score better aligns with human visual perception. A crucial step in the LPIPS metric is the determination of the weights $w_l$. These weights are typically obtained through training, which involves using a set of images and their distorted versions and optimizing the weights based on human subjective evaluations to make the LPIPS score as consistent as possible with human judgments.

These metrics offer means to evaluate the quality of images reconstructed by Multispectral-NeRF model from different perspectives, thereby ensuring the model's high performance and practical application value.

## 5.2 Analysis of Training Results

Based on the experimental area, the real - world 3D reconstruction of the experimental area is carried out using the Multispectral - NeRF model proposed in Section 3, and then it is analyzed according to the model training accuracy evaluation metrics mentioned above. Since the Multispectral - NeRF model proposed in this paper performs 3D model reconstruction based on six - band multispectral images, while currently common models are trained and generated based on RGB true - color photos, and the scope of the experimental area is similar to that included in the large - scale datasets of Mega – NeRF [47] and UE4 – NeRF [48], the metrics of the NeRFacto model [28] and common models trained on the large - scale true - color datasets of Mega – NeRF and UE4 – NeRF are selected for comparison with the metrics obtained by the model proposed in this paper when reconstructing the large - scale multispectral dataset in the experimental area.

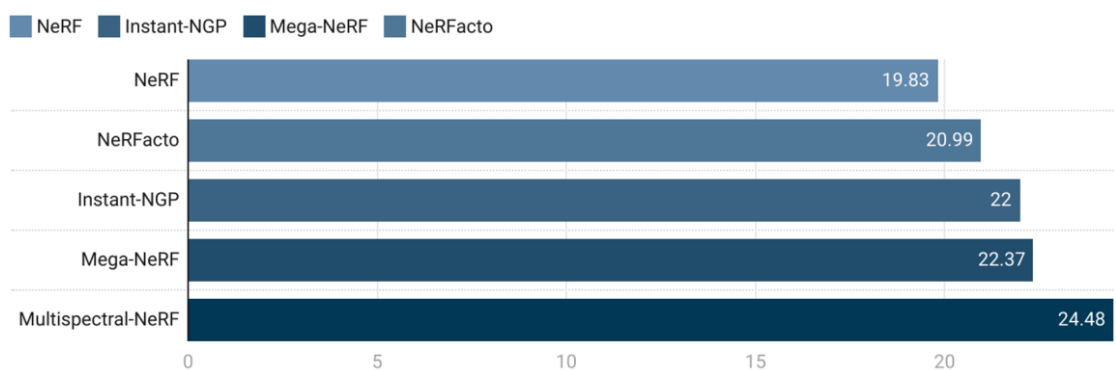

**Figure 14.** PSNR of different models

According to the comparison of model metrics in the experimental region of **Figure 14**, since the NeRFacto model is primarily designed for small-scale unbounded datasets, most NeRF variants tailored for large-scale data—such as Mega-NeRF and UE4-NeRF—achieve PSNR values within the range of 18–25 [49]. Therefore, the proposed model demonstrates superior performance when trained on large-scale scene datasets compared to the original NeRFacto model under similar conditions, achieving performance levels comparable to other models specifically designed for large datasets. For bounded real-world scenes captured by NeRFacto, the reported PSNR values fall within the range of 20–25. Consequently, Multispectral-NeRF shows slightly lower performance in terms of PSNR.

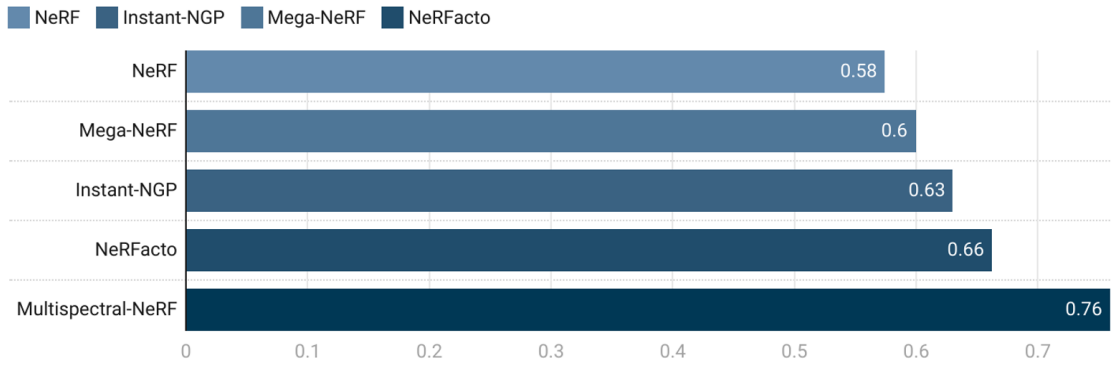

**Figure 15.** SSIM of different models

In terms of SSIM (**Figure 15**), the Multispectral – NeRF model proposed in this paper outperforms common improved NeRF models based on large - scale datasets. The SSIM values of models such as Instant - NGP and Mega - NeRF are approximately in the range of 0.6 to 0.7, while the Multispectral – NeRF model achieves a value above 0.7. This indicates that the reconstruction results of the Multispectral – NeRF model maintain better consistency in the lighting conditions of the real scene during restoration. This advantage stems from the ability to restore lighting conditions from more dimensions by combining multi - spectral and multi - band information. Although the PSNR value of the Multispectral – NeRF is slightly lower than that of the NeRFacto model, it can still effectively capture the structure and visual features of bounded real - world scenes.

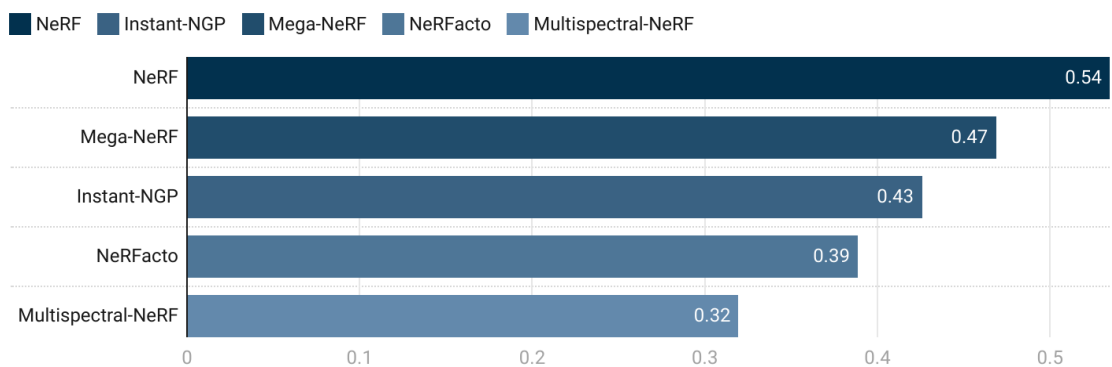

**Figure 16.** LPIPS of different models

On the other hand, the LPIPS (**Figure 16**) values of the Multispectral – NeRF model are generally lower than those of models based on large - scale datasets, such as Instant - NGP and Mega - NeRF, as well as the original NeRFacto model. This compensates for the evaluation limitations of the PSNR and SSIM metrics, which may not fully measure the reconstruction effect in some smooth parts of the image. It indicates that the reconstruction results of the Multispectral – NeRF model are more visually consistent with the real effects perceived by the human eye.

Furthermore, we conducted multispectral 3D reconstruction using both the original NeRFacto model (without parameter modifications) and the modified parameter version.

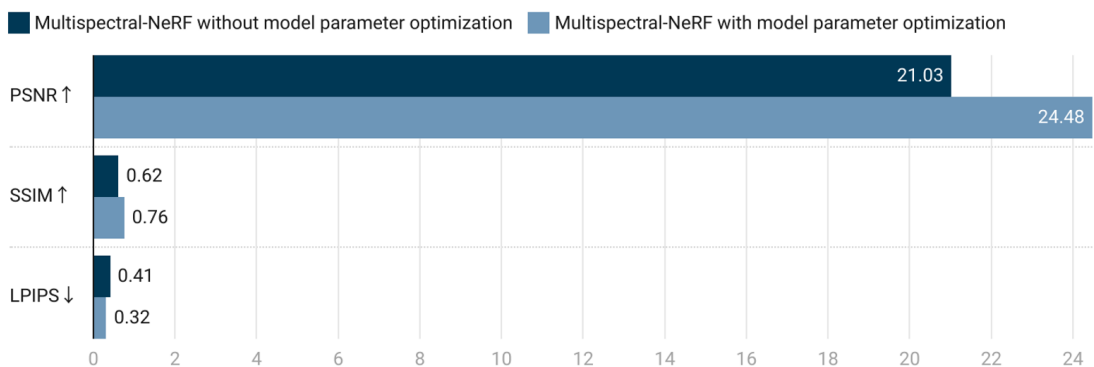

Figure 17. Ablation study of the Multispectral-NeRF model before and after parameter optimization

We ablate the parameter optimization of Multispectral-NeRF, as evidenced in **Figure 17**, although the Multispectral-NeRF model without parameter improvement can incorporate the spectral information of six - band multispectral data, using the model structure and parameters of the original NeRFacto model fails to fully leverage the advantage of the rich spectral information in multispectral data. In terms of reconstruction accuracy, it is still similar to other improved NeRF algorithms. After parameter adjustment to adapt to the characteristics of multispectral data, the Multispectral-NeRF model can read and utilize the spectral information of additional bands beyond the visible light in the training process, preserve the original bit - depth of the multispectral data, and correct the errors between the reconstruction results and the real scene through a higher - dimensional residual function. Ultimately, the reconstruction accuracy of the results is further improved, ensuring that the reconstructed scene can truly reflect the spectral information of the original scene.

In conclusion, the training accuracy of the Multispectral-NeRF model based on oblique multi - spectral photography proposed in this paper has reached the accuracy of NeRFacto on the average dataset, which proves the feasibility and accuracy of the model proposed in this experiment.

# 6. Discussion

In the construction of the Multispectral - NeRF model, considering the hardware resource usage is crucial. In this section, quantitative calculations were performed based on the model's computational parameters related to CPU memory and GPU video memory to ensure optimal performance under different hardware conditions.

## 6.1 CPU Parameter Selection

The Multispectral - NeRF model can learn more detailed and accurate information from a large number of input training photos. However, this requires more CPU memory to store intermediate representations and gradients during training. During the training process of the Multispectral - NeRF model, the CPU is responsible for storing data, including training photos and their pose information, before it is loaded onto the GPU for training. For multispectral images, which consume more memory than regular images, direct storage may lead to issues such as memory overflow. Therefore, a batch - by - batch resampling strategy is required.

The memory occupied during the training of the Multispectral - NeRF model is also related to the resolution of each image. Take an image with a resolution of $H*W$ as an example. In each batch, $N$ images are randomly selected, and $N*H*W$ rays are generated based on each pixel of these images. When resampling images in the $M$-th batch, another $N$ images are randomly selected from the new batch, and again $N*H*W$ rays are generated based on each pixel of these new images. Therefore, the peak CPU usage occurs during the data initialization phase of every $M$ batches. Taking the resolution of 1280*960 in oblique multispectral photography as an example, with 4000 images, the CPU usage at the peak reaches 180 - 200G. Therefore, for the multispectral images used in this paper, the maximum value of N is approximately 1200. This ensures that more images can participate in the training multiple times while maintaining high - speed training to achieve the best results. In addition, parameter selection was carried out on two servers with different configurations in this experiment. The final quantitative selection of the optimal parameters is shown in **Table 8**:

**Table 8.** Parameter selections of Multispectral-NeRF CPU optimal

| CPU configuration | Number of photos | N | M | Max | Memory usage |
|---|---|---|---|---|---|
| 256G | 4000 | 1000 | 2000 | 120000 | 180~240G |
| 512G | 4000 | 2000 | 4000 | 60000 | 340~460G |

## 6.2 GPU Parameter Selection

In the training of Multispectral - NeRF, GPUs are primarily used to store model parameters, gradients, intermediate values during the forward and backward passes of training, and the light rays generated from training photos. GPU memory is a limiting factor for the model size and the number of light rays processed in parallel. Increasing the number of training photos or light rays directly affects the GPU memory usage. Finding a balance among the number of training photos, the number of sampled light rays, and GPU memory ensures that the model can represent the scene in sufficient detail and accuracy while still being trainable within hardware limitations. This paper uses a GeForce RTX 2080Ti graphics card with 11GB of video memory and an NVIDIA A100 graphics card with 80GB of video memory for model calculations.

Multispectral - NeRF controls the GPU usage of the model through the following two key parameters: the number of training rays per batch and the number of evaluation rays per batch. The number of training rays per batch indicates how many rays are selected for the next sampling point selection in each training batch in the NeRF sampling point module. The number of evaluation rays per batch indicates how many rays are selected for the next sampling point selection in each evaluation batch in the NeRF sampling point module. A larger number of rays per batch can more efficiently utilize the parallel computing power of the GPU, accelerate the training speed, and rationally utilize video memory resources. When this paper sets the two parameter values to 32768 and 16384, the GPU usage of the Multispectral - NeRF model with improved parameters is shown in **Table 9**:

**Table 9.** Parameter selections of Multispectral-NeRF training ray batch number

| GPU | Training Ray Batch Size | Evaluation Ray Batch Size | Video memory usage |
|---|---|---|---|
| GeForce RTX 2080Ti(11GB) | 16384 | 8192 | 10G |
| NVIDIA A100 (80GB) | 32768 | 16384 | 16G |

For GPUs with VRAM below 12GB, parameter allocation should be proportionally adjusted based on available memory. Given that our MLP architecture restricts hidden layers to specific dimensions (16, 32, 64, or 128 neural units), the training parameters must exceed the default 4096 threshold while maintaining 128-unit multiples. The baseline configuration (4096 parameters) requires 6GB VRAM during training.

Empirical values of 8092 or 4096 parameters are generally recommended for standard GPUs to balance computational efficiency and memory constraints.

## 6.3 Shortcomings and Prospects

Although the proposed Multispectral-NeRF model achieves 3D reconstruction of multispectral data and demonstrates certain accuracy advantages over existing methods, there are still several limitations to the current study. The multispectral data used in this research consists of only six spectral bands, whereas typical hyperspectral imagery comprises hundreds of bands. By modifying the input architecture of the model, it may be possible to adapt the framework for higher spectral resolution. Therefore, future work could explore the potential of UAV-based hyperspectral 3D reconstruction and investigate whether similar methodologies can be applied to develop 3D hyperspectral models. Furthermore, due to the volumetric rendering mechanism of NeRF, noise artifacts may appear in the reconstructed 3D point cloud, particularly in regions close to the terrain surface. Future research could incorporate point cloud filtering and smoothing techniques to enhance the quality and accuracy of multispectral 3D models.

## 7. Conclusion

This paper addresses the issues of traditional 3D modeling methods in handling multi - spectral data fusion. Based on the research of real - scene 3D reconstruction using oblique multi - spectral photography and considering the complexity of surface spectral characteristics and multi - spectral dimensions, the School of Geographical Sciences at the Xianlin Campus of Nanjing Normal University is selected as the study area. By comprehensively considering the influence of various parameters on the model and estimating the integrity and accuracy of geographical coordinates and spectral information, a Multispectral – NeRF model based on oblique multi - spectral photography is proposed. Quantitative index analysis, evaluation, and error analysis are conducted on the experimental results of each stage of the proposed model to verify the effectiveness of the proposed model and method. The evaluation of model training accuracy shows that the model proposed in this paper can process a wider range of bands and deeper data in multi - spectral data, and then reconstruct the 3D model of the experimental area through implicit representation. In addition, the reconstructed scene can truly reflect the spectral attributes of the original scene, and the evaluation indicators of model training basically reach the level of the original model on the average dataset.

**Author Contributions:**

**Funding:**

**Data Availability Statement:**

**Conflicts of Interest:**

# References

1. Pollefeys, M.; Nistér, D.; Frahm, J.-M.; Akbarzadeh, A.; Mordohai, P.; Clipp, B.; Engels, C.; Gallup, D.; Kim, S.-J.; Merrell, P.; Salmi, C.; Sinha, S.; Talton, B.; Wang, L.; Yang, Q.; Stewénius, H.; Yang, R.; Welch, G.; Towles, H. Detailed Real-Time Urban 3D Reconstruction from Video. *Int J Comput Vis* **2007**, *78* (2–3), 143–167. https://doi.org/10.1007/s11263-007-0086-4.

2. Wu, J.; Wyman, O.; Tang, Y.; Pasini, D.; Wang, W. Multi-View 3D Reconstruction Based on Deep Learning: A Survey and Comparison of Methods. *Neurocomputing* **2024**, *582*, 127553. https://doi.org/10.1016/j.neucom.2024.127553.

3. Lu, Y.; Wang, S.; Fan, S.; Lu, J.; Li, P.; Tang, P. Image-Based 3D Reconstruction for Multi-Scale Civil and Infrastructure Projects: A Review from 2012 to 2022 with New Perspective from Deep Learning Methods. *Advanced Engineering Informatics* **2024**, *59*, 102268. https://doi.org/10.1016/j.aei.2023.102268.

4. Shiode, N. 3D urban models: Recent developments in the digital modelling of urban environments in three-dimensions. *GeoJournal* **2000**, *52* (3), 263–269. https://doi.org/10.1023/a:1014276309416.

5. Fruh, C. and Zakhor, A. (n.d.). 3D model generation for cities using aerial photographs and ground level laser scans. *Proceedings of the 2001 IEEE Computer Society Conference on Computer Vision and Pattern Recognition*. *CVPR* **2001**, *2*, II-31-II–38.

6. Xie, F.; Lin, Z.; Gui, D.; Lin, H. STUDY ON CONSTRUCTION OF 3D BUILDING BASED ON UAV IMAGES. *Int. Arch. Photogramm. Remote Sens. Spatial Inf. Sci.* **2012**, *XXXIX-B1*, 469–473. https://doi.org/10.5194/isprsarchives-xxxix-b1-469-2012.

7. LIU, Z.; GUO, H.; WANG, C. Considerations on Geospatial Big Data. *IOP Conf.*

*Ser.: Earth Environ. Sci.* **2016**, *46*, 012058. https://doi.org/10.1088/1755-1315/46/1/012058.

8. Zainuddin, K.; Majid, Z.; Ariff, M. F. M.; Idris, K. M.; Abbas, M. A.; Darwin, N. 3D MODELING FOR ROCK ART DOCUMENTATION USING LIGHTWEIGHT MULTISPECTRAL CAMERA. *Int. Arch. Photogramm. Remote Sens. Spatial Inf. Sci.* **2019**, *XLII-2/W9*, 787–793. https://doi.org/10.5194/isprs-archives-xlii-2-w9-787-2019.

9. Li, Z.; Schaefer, M.; Strahler, A.; Schaaf, C.; Jupp, D. On the Utilization of Novel Spectral Laser Scanning for Three-Dimensional Classification of Vegetation Elements. *Interface Focus.* **2018**, *8* (2), 20170039. https://doi.org/10.1098/rsfs.2017.0039.

10. Nijland, W.; Coops, N. C.; Nielsen, S. E.; Stenhouse, G. Integrating Optical Satellite Data and Airborne Laser Scanning in Habitat Classification for Wildlife Management. *International Journal of Applied Earth Observation and Geoinformation* **2015**, *38*, 242–250. https://doi.org/10.1016/j.jag.2014.12.004.

11. Over, J.-S. R., *et al.* Processing coastal imagery with Agisoft Metashape Professional Edition, version 1.6—Structure from motion workflow documentation. In Open-File Report. US Geological Survey **2021**.

12. Wan, Q., *et al.* Constraining the Geometry of NeRFs for Accurate DSM Generation from Multi-View Satellite Images. *ISPRS International Journal of Geo-Information*, **2024**, *13*(7), 243.

13. Johari, M. M.; Lepoittevin, Y.; Fleuret, F. GeoNeRF: Generalizing NeRF with Geometry Priors. arXiv **2021**. https://doi.org/10.48550/ARXIV.2111.13539.

14. Mokssit, S.; Licea, D. B.; Guermah, B.; Ghogho, M. Deep Learning Techniques for Visual SLAM: A Survey. *IEEE Access* **2023**, *11*, 20026–20050. https://doi.org/10.1109/access.2023.3249661.

15. Li, Y.; Ma, L.; Zhong, Z.; Liu, F.; Chapman, M. A.; Cao, D.; Li, J. Deep Learning for LiDAR Point Clouds in Autonomous Driving: A Review. *IEEE Trans. Neural Netw. Learning Syst.* **2021**, *32* (8), 3412–3432. https://doi.org/10.1109/tnnls.2020.3015992.

16. Delegido, J.; Verrelst, J.; Meza, C. M.; Rivera, J. P.; Alonso, L.; Moreno, J. A Red-Edge Spectral Index for Remote Sensing Estimation of Green LAI over Agroecosystems. *European Journal of Agronomy* **2013**, *46*, 42–52. https://doi.org/10.1016/j.eja.2012.12.001.

17. Zhang, Y.; Migliavacca, M.; Penuelas, J.; Ju, W. Advances in Hyperspectral Remote Sensing of Vegetation Traits and Functions. *Remote Sensing of Environment* **2021**, *252*, 112121. https://doi.org/10.1016/j.rse.2020.112121.

18. Jonathan A.R. Rall, and R. G. K. (n.d.). Spectral ratio biospheric Lidar. *IEEE International IEEE International IEEE International Geoscience and Remote Sensing Symposium, 2004. IGARSS '04. Proceedings*. **2004**, *3*, 1951–1954.

19. Decker, K. T.; Borghetti, B. J. Composite Style Pixel and Point Convolution-Based Deep Fusion Neural Network Architecture for the Semantic Segmentation of Hyperspectral and Lidar Data. *Remote Sensing* **2022**, *14* (9), 2113. https://doi.org/10.3390/rs14092113.

20. Vlaminck, M.; Diels, L.; Philips, W.; Maes, W.; Heim, R.; Wit, B. D.; Luong, H. A Multisensor UAV Payload and Processing Pipeline for Generating Multispectral Point Clouds. *Remote Sensing* **2023**, *15* (6), 1524. https://doi.org/10.3390/rs15061524.

21. Murtiyoso, *et al.* A NOVEL AND AUTOMATIC PHOTOGRAMMETRIC WORKFLOW FOR 3D POINT CLOUD AI-BASED SEMANTIC SEGMENTATION. *International Symposium on Applied Geoinformatics* **2021**.

22. Karantanellis, E.; Arav, R.; Dille, A.; Lippl, S.; Marsy, G.; Torresani, L.; Oude Elberink, S. EVALUATING THE QUALITY OF PHOTOGRAMMETRIC POINT-CLOUDS IN CHALLENGING GEO-ENVIRONMENTS – A CASE STUDY IN AN ALPINE VALLEY. *Int. Arch. Photogramm. Remote Sens. Spatial Inf. Sci.* **2020**, *XLIII-B2-2020*, 1099–1105. https://doi.org/10.5194/isprs-archives-xliii-b2-2020-1099-2020.

23. Mildenhall, B., Srinivasan, P. P., Tancik, M., Barron, J. T., Ramamoorthi, R., & Ng, R. NeRF: Representing Scenes as Neural Radiance Fields for View Synthesis. In *Lecture Notes in Computer Science* (pp. 405–421). Springer International Publishing. arXiv **2020**. https://doi.org/10.1007/978-3-030-58452-8_24

24. Yang, B.; Rosa, S.; Markham, A.; Trigoni, N.; Wen, H. Dense 3D Object Reconstruction from a Single Depth View. *arXiv* **2018**. https://doi.org/10.48550/ARXIV.1802.00411.

25. Tatarchenko, M.; Dosovitskiy, A.; Brox, T. Octree Generating Networks: Efficient Convolutional Architectures for High-Resolution 3D Outputs. arXiv **2017**. https://doi.org/10.48550/ARXIV.1703.09438.

26. Yagubbayli, F.; Wang, Y.; Tonioni, A.; Tombari, F. LegoFormer: Transformers for

Block-by-Block Multi-View 3D Reconstruction. arXiv **2021**. https://doi.org/10.48550/ARXIV.2106.12102.

27. Mandikal, P.; Navaneet, K. L.; Agarwal, M.; Babu, R. V. 3D-LMNet: Latent Embedding Matching for Accurate and Diverse 3D Point Cloud Reconstruction from a Single Image. arXiv 2018. https://doi.org/10.48550/ARXIV.1807.07796.

28. Tancik, M.; Weber, E.; Ng, E.; Li, R.; Yi, B.; Kerr, J.; Wang, T.; Kristoffersen, A.; Austin, J.; Salahi, K.; Ahuja, A.; McAllister, D.; Kanazawa, A. Nerfstudio: A Modular Framework for Neural Radiance Field Development. *arXiv* **2023**. https://doi.org/10.48550/ARXIV.2302.04264.

29. Verbin, D.; Hedman, P.; Mildenhall, B.; Zickler, T.; Barron, J. T.; Srinivasan, P. P. Ref-NeRF: Structured View-Dependent Appearance for Neural Radiance Fields. *IEEE Trans. Pattern Anal. Mach. Intell.* **2024**, 1–12. https://doi.org/10.1109/tpami.2024.3360018.

30. Müller, T.; Evans, A.; Schied, C.; Keller, A. Instant Neural Graphics Primitives with a Multiresolution Hash Encoding. *arXiv* **2022**. https://doi.org/10.48550/ARXIV.2201.05989.

31. Chen, A.; Xu, Z.; Geiger, A.; Yu, J.; Su, H. TensoRF: Tensorial Radiance Fields. arXiv **2022**. https://doi.org/10.48550/ARXIV.2203.09517

32. Wang, Z.; Wu, S.; Xie, W.; Chen, M.; Prisacariu, V. A. NeRF--: Neural Radiance Fields Without Known Camera Parameters. arXiv **2021**. https://doi.org/10.48550/ARXIV.2102.07064.

33. Lin, C.-H.; Ma, W.-C.; Torralba, A.; Lucey, S. BARF: Bundle-Adjusting Neural Radiance Fields. In *2021 IEEE/CVF International Conference on Computer Vision (ICCV)*; IEEE, **2021**; pp 5721–5731. https://doi.org/10.1109/iccv48922.2021.00569.

34. Li, J., Li, Y., Sun, C., Wang, C., & Xiang, J. Spec-NeRF: Multi-spectral Neural Radiance Fields (Version 1). arXiv **2023**. https://doi.org/10.48550/ARXIV.2310.12987

35. Poggi, M., Ramirez, P. Z., Tosi, F., Salti, S., Mattoccia, S., & Di Stefano, L. (2022, September). Cross-spectral neural radiance fields. In *2022 International Conference on 3D Vision (3DV)* (pp. 606-616). IEEE.

36. Tulsiani, S.; Zhou, T.; Efros, A. A.; Malik, J. Multi-View Supervision for Single-View Reconstruction via Differentiable Ray Consistency. arXiv **2017**. https://doi.org/10.48550/ARXIV.1704.06254.

37. Kerbl, B., Kopanas, G., Leimkuehler, T., & Drettakis, G. 3D Gaussian Splatting for Real-Time Radiance Field Rendering. *ACM Transactions on Graphics*, *42*(4), 1–14. arXiv **2023**. https://doi.org/10.1145/3592433
38. Xie, Y., Takikawa, T., Saito, S., Litany, O., Yan, S., Khan, N., Tombari, F., Tompkin, J., sitzmann, V., & Sridhar, S. Neural Fields in Visual Computing and Beyond. Computer Graphics Forum, 41(2), 641–676. arXiv **2022**. https://doi.org/10.1111/cgf.14505
39. **ao, W., Chierchia, R., Cruz, R. S., Li, X., Ahmedt-Aristizabal, D., Salvado, O., ... & Lebrat, L. Neural Radiance Fields for the Real World: A Survey. *arxiv preprint arxiv:2501.13104*.
40. Cai, J., & Lu, H. (2024, May). Nerf-based multi-view synthesis techniques: A survey. In *2024 International Wireless Communications and Mobile Computing (IWCMC)* (pp. 208-213). IEEE.
41. Wu, T., Yuan, Y.-J., Zhang, L.-X., Yang, J., Cao, Y.-P., Yan, L.-Q., & Gao, L. Recent advances in 3D Gaussian splatting. *Computational Visual Media*, *10*(4), 613–642. arXiv **2024**. https://doi.org/10.1007/s41095-024-0436-y
42. Hu, R., He, Q., Du, D., & Jin, X. ScatterSplatting: Enhanced View Synthesis in Scattering Scenarios via Joint NeRF and Gaussian Splatting. *2025 IEEE International Symposium on Circuits and Systems (ISCAS)*, 1–5. https://doi.org/10.1109/iscas56072.2025.11044088
43. Furukawa, Y., & Ponce, J. Accurate, Dense, and Robust Multiview Stereopsis. *IEEE Transactions on Pattern Analysis and Machine Intelligence*, *32*(8), 1362–1376. arXiv **2010**. https://doi.org/10.1109/tpami.2009.161
44. Ramamoorthi, R., & Hanrahan, P. An efficient representation for irradiance environment maps. *Proceedings of the 28th Annual Conference on Computer Graphics and Interactive Techniques*, 497–500. arXiv **2001**. https://doi.org/10.1145/383259.383317
45. Zhang, J., Wang, X., Ni, X., Dong, F., Tang, L., Sun, J., & Wang, Y. Neural radiance fields for multi-scale constraint-free 3D reconstruction and rendering in orchard scenes. *Computers and Electronics in Agriculture*, *217*, 108629. arXiv **2024**. https://doi.org/10.1016/j.compag.2024.108629
46. Fundamental geographic information — specifications for the producing of three-dimensional model. *CH/T 9016-2012*.
47. Turki, H., Ramanan, D., & Satyanarayanan, M. *Mega-NeRF: Scalable Construction of Large-Scale NeRFs for Virtual Fly-Throughs* (Version 2). arXiv **2021**. https://doi.org/10.48550/ARXIV.2112.10703
48. Gu, J., Jiang, M., Li, H., Lu, X., Zhu, G., Shah, S. A. A., Zhang, L., & Bennamoun,

M. *UE4-NeRF:Neural Radiance Field for Real-Time Rendering of Large-Scale Scene* (Version 1). arXiv **2023**. https://doi.org/10.48550/ARXIV.2310.13263

49. Xu L, Xiangli Y, Peng S, Pan X, Zhao N, Theobalt C, Dai B, Lin D. Grid-guided neural radiance fields for large urban scenes. In *Proceedings of the IEEE/CVF Conference on Computer Vision and Pattern Recognition*, **2023**(pp. 8296-8306).